\documentclass{article}

\usepackage{PRIMEarxiv}
\usepackage[utf8]{inputenc} 
\usepackage[T1]{fontenc}    
\usepackage{url}            
\usepackage{booktabs}       
\usepackage{amsfonts}       
\usepackage{nicefrac}       
\usepackage{microtype}      
\usepackage{lipsum}
\usepackage{fancyhdr}       
\usepackage{amsmath}
\usepackage{cite}
\usepackage{amsfonts}
\usepackage{color}
\usepackage{graphicx}       
\usepackage{hyperref}       
\usepackage[toc,page]{appendix}

\begin{document}
\pagestyle{fancy}
\thispagestyle{empty}
\rhead{ \textit{ }} 

\fancyhead[LO]{SAMAug: Point Prompt Augmentation for Segment Anything Model}

\title{SAMAug: Point Prompt Augmentation for Segment Anything Model}

\author{
Haixing Dai\textsuperscript{$\dagger$},
Chong Ma\textsuperscript{$\dagger$},
Zhiling Yan,
Zhengliang Liu,
Enze Shi,
Yiwei Li,\\
\textbf{Peng Shu,
Xiaozheng Wei,
Lin Zhao,
Zihao Wu, 
Fang Zeng,}\\
\textbf{Dajiang Zhu,
Wei Liu,
Quanzheng Li,
Lichao Sun,
Shu Zhang, 
Tianming Liu,}
and \textbf{Xiang Li}
\thanks{
Corresponding author: Xiang Li (email: xli60@mgh.harvard.edu) \\
\textsuperscript{{$\dagger$}} These authors contributed equally to this paper. \\
Haixing Dai, Zhengliang Liu, Yiwei Li, Lin Zhao, Zihao Wu, and Tianming Liu are with the School of Computing, University of Georgia, Athens, GA, USA. (e-mail: hd54134@uga.edu; zl18864@uga.edu; yl80817@uga.edu; lin.zhao@uga.edu; zw63397@uga.edu; tliu@uga.edu) \\
Chong Ma and Xiaozheng Wei are with the School of Automation, Northwestern Polytechnical University, Xi’an, China. (e-mail:  mcnpu@mail.nwpu.edu.cn; wxz41@mail.nwpu.edu.cn) \\
Zhiling Yan and Lichao Sun are with the Department of Computer Science and Engineering,  Lehigh University, Bethlehem, PA, USA. (e-mail:  zhy423@lehigh.edu; lis221@lehigh.edu) \\
Fang Zeng, Quanzheng Li, and Xiang Li are with the Department of Radiology, Massachusetts General Hospital and Harvard Medical School, Boston, MA, USA. (e-mail: fzeng1@mgh.harvard.edu; li.quanzheng@mgh.harvard.edu; xli60@mgh.harvard.edu) \\
Dajiang Zhu is with the Department of Computer Science and Engineering, University of Texas at Arlington, TX, USA.(e-mail: dajiang.zhu@uta.edu) \\
Wei Liu is with the Department of Radiation Oncology, Mayo Clinic, USA. (e-mail: liu.wei@mayo.edu) \\
Shu Zhang is with the School of Computer Science, Northwestern Polytechnical University, Xi’an, China. (e-mail: 
shu.zhang@nwpu.edu.cn)
}}

\maketitle

\begin{abstract}
This paper introduces SAMAug, a novel visual point augmentation method for the Segment Anything Model (SAM) that enhances interactive image segmentation performance. SAMAug generates augmented point prompts to provide more information about the user's intention to SAM. Starting with an initial point prompt, SAM produces an initial mask, which is then fed into our proposed SAMAug to generate augmented point prompts. By incorporating these extra points, SAM can generate augmented segmentation masks based on both the augmented point prompts and the initial prompt, resulting in improved segmentation performance. We conducted evaluations using four different point augmentation strategies: random sampling, sampling based on maximum difference entropy, maximum distance, and saliency. Experiment results on the COCO, Fundus, COVID QU-Ex, and ISIC2018 datasets show that SAMAug can boost SAM's segmentation results, especially using the maximum distance and saliency. SAMAug demonstrates the potential of visual prompt augmentation for computer vision. Codes of SAMAug are available at \href{https://github.com/yhydhx/SAMAug}{github.com/yhydhx/SAMAug}.
\end{abstract}


\section{Introduction} \label{Introduction}
Recent advancement in large pre-trained models \cite{liu2023summary,zhao2023brain} inspired a series of developments of foundational models in computer vision \cite{willemink2022toward,zhou2023comprehensive}. Among these, the Segment Anything Model (SAM) \cite{kirillov2023segment} stands out as a novel interactive model specifically designed for image segmentation tasks and subsequent downstream applications.

SAM represents the paradigm shift in artificial intelligence towards large-scale, general-purposed "foundation" models \cite{yuan2023power,zhang2023biomedgpt}. The SAM model demonstrates the feasibility of performing multi-task image segmentation in a zero-shot learning approach. It also allows the use of "prompts" in the form of user-provided points, bounding boxes, or dense masks to perform the segmentation via an in-context-learning scheme. However, simple prompts such as a single point will lead to ambiguities in the SAM model, which is also discussed in the original paper of SAM  \cite{kirillov2023segment}. While the multi-mask prediction and confidence score-based result selection scheme adopted by SAM can solve some of the ambiguities, we observed that it is also possible to generate additional point prompts automatically to improve the performance of SAM. In this work, we propose SAMAug, a novel visual point augmentation method for generating additional segmentation masks using SAM. SAMAug performs prompt (point or bounding-box) augmentation based on the initial segmentation results from SAM using a single point or box prompt, then incorporates them into the prompt-based segmentation by SAM. We evaluate four sampling approaches for point prompts: random, maximum difference entropy, maximum distance, and saliency-based. We also investigated the inner- and outer-bounding box augmentation for box prompts. 

With the experiments on the COCO, Fundus, COVIDQU-Ex, and ISIC2018 datasets, we demonstrate that SAMAug can improve SAM's performance, especially when using the maximum distance and saliency model point selection approaches. Our results showcase more possibilities of incorporating human inputs in a visual prompt setting. SAMAug represents an important step towards prompt-based augmentation methods for computer vision that can reduce user inputs and improve model performance.

This study makes the following key contributions:
\begin{itemize}
    \item We develop a novel visual point augmentation framework, SAMAug, for generating additional prompts without extra manual operations for SAM.
    \item We propose a visual prompt augmentation theory based on the invariance in prompt selection.
    \item We tested different techniques for visual prompt augmentation with experiments on one general-domain and three medical-domain datasets and identified the most effective augmentation techniques (maximum distance and saliency).
\end{itemize}
   
\section{Background and Related work}
\subsection{Segment Anything Model}
Recently, the introduction of the Segment Anything Model (SAM) has revolutionized the approach to image segmentation. It is a general, prompt-able model designed to adapt to specific tasks via an in-context learning approach, much like the text prompts used in natural language processing models \cite{brown2020language,liu2023summary}. Segmentation models can be categorized into two broad classes \cite{zhang2023comprehensive}: Interactive segmentation requires user input to refine a mask iteratively, whereas automatic segmentation requires a significant amount of manually annotated objects for training. SAM \cite{kirillov2023segment} unifies these two classes of approaches. It is a single model capable of performing both interactive and automatic segmentation. Its interface is designed to handle a wide array of segmentation tasks, facilitated by an appropriate prompt for the model. One of SAM's most distinguishing features is its training on an unprecedentedly large dataset of over one billion annotations collected as part of the Segment Anything project. This diverse, high-quality dataset enables SAM to generalize to new types of objects and images beyond what it observed during training.

\subsection{Research Work based on SAM Model}
SAM is a very powerful fundamental model for image segmentation that can achieve zero-short learning on unseen domains, including medical image analysis \cite{li2023artificial}. There are many works verifying the effect of SAM for medical tasks, such as \cite{zhang2023segment}, \cite{deng2023segment}, and \cite{he2023accuracy}. It is found that even without any further training or transfer learning of the model, SAM can achieve good segmentation performance. Work in \cite{zhang2023input} leverages SAM for generating samples that can be used to train other medical image segmentation models. Further development of domain-specific SAM with model fine-tuning \cite{wu2023medical,chen2023ma,kim2023medivista} shows that SAM with adaptation to the new dataset can achieve state-of-the-art performance over a wide spectrum of tasks. Still, SAM will fail when the segmentation target is small, dense, or irregular. Experiments also show that the segmentation quality of SAM can be improved by adjusting prompts \cite{huang2023segment}, \cite{wu2023medical}. Therefore, exploring prompt tuning can be a solution for improving its performance in the medical domain.

\subsection{Prompt Learning and Augmentation}
Prompt learning can extract valuable insights from large pre-trained models \cite{diao2022black,wang2023prompt}. This technique revolves around optimizing either a sequence of tokens (discrete prompts) \cite{gao2020making,ben2022pada} or a sequence of vectors (continuous prompts) \cite{qin2021learning,liu2021gpt}. Prompt-based learning allows users to effectively exploit the potential of large pre-trained models without the need for updating model parameters (e.g., via fine-tuning). It is also possible and potentially beneficial to augment prompts \cite{shum2023automatic}. This strategy, distinct from data augmentation, involves the generation and optimization of prompts, thereby further harnessing the capabilities of prompt-based models \cite{dai2023chataug}. One prompt augmentation method is AutomateCOT \cite{shum2023automatic}, which automatically augments and selects prompts to enhance the reasoning ability of LLMs. AutomateCoT generates pseudo-chains for input questions and then prunes incorrect ones based on the predicted answers. It finally selects an optimal combination of exemplars using a variance-reduced policy gradient strategy. The proposed SAMAug shares a similar motivation with AutomateCoT while focusing on the computer vision tasks by augmenting visual prompts for SAM, a large vision model, to produce augmented segmentation masks. In contrast, AutomateCoT generates additional textual prompts for language models to solve NLP tasks such as arithmetic and symbolic reasoning. Despite the differences in modality and tasks, both SAMAug and AutomateCoT demonstrate the value of harnessing model capabilities through refined prompt design.

\subsection{Visual Prompts and Interactive Image Segmentation}
By incorporating user input, integrative image segmentation can obtain more precise and targeted results. With clicks (points), strokes (lines), or bounding boxes, human users can provide visual cues to guide the model about areas of interest or to correct potential errors \cite{mcguinness2010comparative}. With the user inputs, the model can then quickly adapt and provide immediate updates, allowing the user to assess and refine the segmentation results \cite{li2015interactive}.
Inspired by the success of in-context learning in NLP, researchers have explored visual prompts in computer vision in a more systematic approach, integrating the visual cues into model learning \cite{xiao2023instruction}. For example, VPT \cite{jia2022visual} introduces a small part of trainable parameters into input space and keeps the entire pre-trained Transformer backbone frozen. By training these parameters regarded as visual prompts, VPT reaches better performance for diverse visual tasks. Convpass \cite{jie2022convolutional} combines vision transformer (ViT) with convolution bypass prompts to mitigate the computational stress. ViPT \cite{zhu2023visual} implements a prompt-tuning method to utilize prior knowledge to learn more information and structures from limited data. In summary, visual prompts enhance the ability of vision models to understand the task (user's intention) and the image.

\subsection{Sampling Methods for Image Analysis}

Sampling is a fundamental problem in the field of statistics \cite{berndt2020sampling} and deep learning \cite{wu2017sampling}, and is classified into two main categories: Probability Sampling and Non-Probability Sampling. Since each sample has an equal chance of being selected, probability sampling is generally used and statistically more likely to choose a sample that is representative of the total. Probability sampling is further divided into the following four groups: (1) Simple random sampling, (2) Stratified sampling, (3)Cluster sampling, and (4) Systematic sampling. Simple random sampling is the most commonly employed sampling technique that gathers a random selection from the entire population, with each unit having an equal chance of selection. The other sampling techniques, stratified sampling, systematic sampling, and cluster sampling, involve the division of the population into subsets or clusters based on the attributes of the samples, followed by sampling from these subsets. These sampling strategies have significantly improved the understanding of the overall population. 
Within the field of computer vision, a single pixel or a small region of an image can be regarded as a sample unit. Sampling strategies that are based on the properties of these sample units are widely employed in various topics, including image alignment \cite{lowe2004distinctive,bay2006surf,ma2019lmr}, image segmentation \cite{ozden2005image,zhao2015image,long2015fully}, and saliency prediction \cite{itti1998model,achanta2008salient,tong2015salient}. Specifically, a common approach for image alignment involves sampling key points and extracting image features from those points for the alignment. By sampling pixels based on the importance of the semantic information they contain, salient information in an image can be identified, forming the basis of saliency detection. Furthermore, pixels that exhibit high levels of saliency in an image are closely associated with their semantic information \cite{ma2023rectify}. Thus, saliency is often considered an essential criterion when selecting sample points.

\begin{figure*}[ht]
\centering
\includegraphics[width=0.95\textwidth]{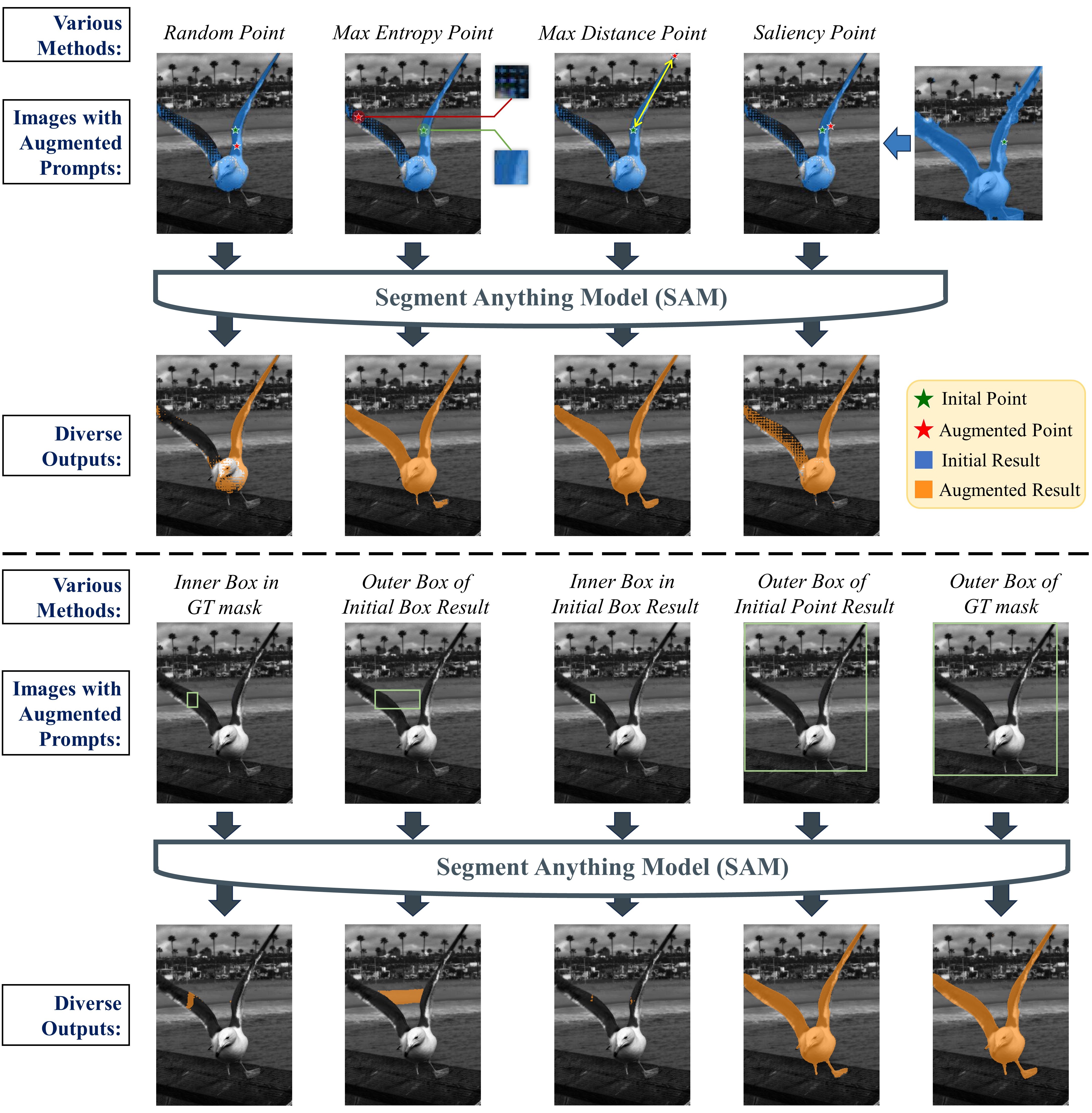}
\caption{The framework of our SAMAug model.}
\label{fig:SAMAug_framework}
\end{figure*}

\section{Methods} \label{method}

\subsection{Premise for Point Prompt Augmentation}
The decoder of SAM employs prompt-based self-attention and cross-attention, allowing attention to flow both from the encoded prompt to the image and vice versa to update the encoded image and prompt features. This mechanism makes SAM potentially sensitive to the location and number of point prompts, as demonstrated in the right panel of Figure~\ref{fig:SAMAug_framework}. Firstly, the bi-directional cross-attention module within the decoding path heavily relies on the coordinate of the point prompt(s) represented by the positional encoding(s). As the image embeddings will also be updated accordingly, point prompts from different locations, even with similar semantic contexts, will potentially lead to differences in the final results. Secondly, as SAM is trained for performing general segmentation rather than specific tasks, it could not accurately deal with (whether suppress or enhance) segmentation boundaries, especially when the prompt information is limited. Thirdly, as pointed out in the SAM documentation~\cite{kirillov2023segment}, a single prompt such as only one point will cause the segmentation ambiguity issue where the prompt can correspond to multiple valid masks, and the SAM model cannot differentiate which mask the prompt is actually referring to. While the SAM model has adopted an ambiguity-resolving module to generate multiple segmentation masks and rank them based on confidence scores, using multiple prompts will certainly address the issue. 

Thus, in this work, we propose the point prompt augmentation scheme based on the premise that: 1) There exists invariance in the visual prompt selection process by the human user, where the selected prompt is from only one of the many possible locations of the user's prior knowledge about the image. Specifically, similar to the rotation- or shift-invariance which is expected in a classic image processing setting \cite{alvarez1993axioms}, we also expect SAM will produce the same segmentation results based on the manifestation of our intention in the form of prompts, regardless of where exactly the point prompt locates at. 2) As experimental results have demonstrated that SAM cannot achieve such invariance, we will need to perform prompt augmentations to guide the model to better understand our intention for the segmentation. The most intuitive approach will be the automatic generation of additional prompts sampled from the initial segmentation results by SAM using a single human-provided prompt. The sampling-from-mask approach essentially regards the initial segmentation mask as a trustworthy yet potentially incomplete result compared with our target; and aims to improve the results by leveraging the prompt selection invariance and adding the extra point prompts. These automatically generated point prompts can be sampled via specific strategies, which are described below.

\subsection{Point Prompt Augmentation by Random Sampling}
\label{aug_random}
In the random selection method, we will add one point to the initial mask by randomly selecting one point from the candidate points. These candidate points are determined based on the initial segmentation mask generated by SAM. Given a set of candidate points \( C \) derived from the initial mask, the selected point \( p \) is:
\begin{equation}
p = \text{random}(C)
\end{equation}
\subsection{Point Prompt Augmentation by Max Entropy Criterion}
In the max entropy point method, we will select a point that maximizes the difference in entropy with respect to the initial prompt. To calculate the entropy, we use a 9x9 grid centered at each candidate point. The entropy of each candidate point is computed based on the distribution of pixel intensities within this grid. The point with the maximum difference in entropy, compared to the initial point, is chosen for addition to the mask. We can denote the pixel intensity distribution within a 9x9 grid centered at a candidate point \( p_i \) by \( P(p_i) \). The entropy \( H(p_i) \) of this distribution is:
\begin{equation}
H(p_i) = -\sum_{j} P(p_i)_j \log P(p_i)_j
\end{equation}
Where \( j \) represents all pixel intensities in the distribution. The selected point \( p_{\text{max}} \) is the one that maximizes the difference in entropy with respect to the initial point \( p_0 \):
\begin{equation}
p_{\text{max}} = \arg\max_{p_i \in C} \left( H(p_i) - H(p_0) \right)
\end{equation}

\subsection{Point Prompt Augmentation by Max Distance Criterion}
In the distance-based method, we search for a point that is at a minimal distance from the initial point, measured by metrics such as Euclidean distance. Specifically, given a distance metric \( d(., .) \), the selected point \( p_{\text{max}} \) is:
\begin{equation}
p_{\text{max}} = \arg\max_{p_i \in C} d(p_0, p_i)
\end{equation}

\subsection{Point Prompt Augmentation by the Saliency Map}
\label{aug_saliency}

In the "Saliency" sampling strategy, we implemented a visual Saliency Transformer (VST) \cite{liu2021visual} used in salient object detection (SOD) to perform point prompt augmentation. VST can extract a saliency mask for objects that are visually prominent for a given image. 
To minimize the loss of edge information, we cropped a region based on the initial SAM result and expanded the boundaries by 10 pixels outward.
Subsequently, we feed this cropped image into the VST model to obtain our saliency mask. In this study, we only focus on VST for RGB-SOD. The specific calculation formula is shown as follows, where $I$ represents the cropped image, $VST(I)$ represents the detected saliency mask generated by the VST model, and $p_s$ denotes our augmented point randomly selected from the region of the saliency mask.

\begin{equation}
p_s = \text{random}(VST(I))
\end{equation}

\subsection{Study on Box Prompt}
As SAM allows box-like visual prompts in addition to point prompts, we also examined the SAMAug performance using box-based prompts. Specifically, we considered the following scenarios:
\subsubsection{Inner Box of GT Mask}
\label{In_box_gt}
In this approach, a point is randomly selected within the ground-truth mask as the center of the box, and it is expanded outward until the boundary of the box aligns with the mask boundary. This is illustrated in Figure~\ref{fig:SAMAug_framework} as "Inner Box in GT mask." We refer to this result as the Initial Box Result.

\subsubsection{Outer Box of GT mask}
This approach generates a box that fully contains the GT mask, as depicted in Figure~\ref{fig:SAMAug_framework} as "Outer Box of GT mask." This box prompt contains all the information and boundary features of the object to be segmented.

\subsubsection{Inner Box of Initial Box Result}
This approach selects a box within the previous segmentation mask (rather than GT) as the prompt. The method for generating the box is consistent with "Inner Box in GT Mask".

\subsubsection{Outer Box of Initial Box Result}
\label{Outer_of_initial}
This approach generates a box outside the boundary of the previous segmentation mask, as demonstrated in Figure~\ref{fig:SAMAug_framework} as "Outer Box of Initial Box Result".

\subsubsection{Outer Box of Initial Point Result}
\label{Ou_box_in}
In this approach, we will generate a bounding box containing the initial segmentation result using SAM, this time with a single point as the prompt. In other words, the final prompts to SAM will be one point plus one bounding box. 

\subsection{Dataset}
\subsubsection{COCO Dataset}
The COCO dataset~\cite{cocodataset} is notable for its size and diversity. It contains more than 200,000 images and over 80 object categories, including common objects like people, animals, vehicles, and household items. The dataset is designed to capture objects in realistic contexts, making it suitable for training and evaluating models that need to understand objects in complex scenes. In addition to object labels, the COCO dataset also includes pixel-level segmentation masks for object instances. In our experiment, we randomly selected 20 images in each category, to a total of 1600 images for model evaluation.

\subsubsection{Diabetic Retinopathy Detection Challenge (Fundus) Dataset}
The Diabetic Retinopathy Detection Challenge dataset~\cite{eyepacs_2012} is a collection of 88,702 images of the human fundus, which is the interior surface of the eye opposite the lens. These images are obtained through fundus photography, where a specialized camera captures detailed images of the retina, blood vessels, and other structures within the eye. These datasets include images from both healthy individuals and patients with specific eye conditions or diseases like diabetic retinopathy, macular degeneration, or glaucoma. In our experiment, we used a randomly selected 300-image subset for evaluating SAMAug. 

\subsubsection{ISIC (Skin) Dataset}
We utilized the dataset published by the largest skin image analysis challenge in the world, hosted by the International Skin Imaging Collaboration in 2018 (ISIC 2018)~\cite{codella2019skin}. The dataset for the segmentation task includes a total of 2,594 training images, 100 validation images, and 1,000 test images. In our experiment, we used the 1000-image test set for evaluating SAMAug, and each image is accompanied by its corresponding ground truth mask for lesion delineation.

\subsubsection{COVID QU-Ex (CXR) Dataset}
COVID QU-Ex~\cite{tahir3122958covid} is the most extensive lung mask dataset with pixel-level annotations to detect, localize, and quantify COVID-19 infection from X-ray images. This dataset encompasses a total of 33,920 chest X-ray (CXR) images, including 1) 11,956 COVID-19 cases, 2) 11,263 cases of Non-COVID infections (Viral or Bacterial Pneumonia), and 3) 10,701 normal cases. Comprehensive and accurate lung segmentation masks are provided for the entirety of the dataset. We used the test set consisting of 583 chest X-ray images from the COVID-19 subset as our testing data. Our objective was to utilize the SAM to segment the lung regions within the images and subsequently compare them to the ground truth annotations associated with the images.

\subsection{Implementation Details}
The goal of this work is to perform visual point augmentation in a segmentation task. Model input consists of the initial point prompt and the original image. We adopt the Segmentation Anything Model (SAM) to obtain the initial mask. After obtaining the initial mask from SAM, we sampled additional point prompts using the methods described in the Method section. Specifically, we apply four different methods: random selection, maximum difference entropy, max distance method, and saliency-based. These methods allow us to augment the initial point prompt by selecting spatially or semantically related points as extra point prompts in the next iteration of segmentation. By incorporating the additional point prompts into the segmentation process, SAM will generate new segmentation results, incorporating the information from both the initial and augmented point prompts. The experiments were conducted on an A100 graphics card within the PyTorch environment. The COCO dataset (1600 images) took approximately 4 hours to complete, while the Fundus (300 images), Chest X-ray (583 images), and ISIC2018 (1000 images) datasets required approximately 20 minutes to finish. 

\section{Results}

\begin{table*}[!t]
\caption{Performance Comparison of Point Prompt on COCO, Fundus, COVID QU-Ex, and ISIC2018 datasets. (Unit: Dice score) }
\centering
\label{tab:SAM_Med_Point}
\resizebox{\linewidth}{!}{
\begin{tabular}{lcccccccc}
\hline
Dataset   & Initial       & Random        & Max Entropy   & Max Distance  & Saliency  & GT Random     & GT Max Entropy & GT Max Distance \\
\hline
\noalign{\smallskip}
COCO   &   0.601\(\pm\)0.002     & 0.614\(\pm\)0.002       & 0.621\(\pm\)0.004            & 0.651\(\pm\)0.002  & 0.631\(\pm\)0.001  & 0.794\(\pm\)0.005 & 0.781\(\pm\)0.004  & 0.797\(\pm\)0.007 \\
Fundus & 0.766\(\pm\)0.008  & 0.794\(\pm\)0.007    & 0.791\(\pm\)0.006   & 0.802\(\pm\)0.007   & 0.792\(\pm\)0.002  & 0.840\(\pm\)0.008 & 0.796\(\pm\)0.010  & 0.849\(\pm\)0.010     \\
COVID QU-Ex    & 0.488\(\pm\)0.003 & 0.503\(\pm\)0.003 & 0.490\(\pm\)0.003 & 0.497\(\pm\)0.002 & 0.495\(\pm\)0.002  & 0.556\(\pm\)0.001 & 0.526\(\pm\)0.002  & 0.454\(\pm\)0.004          \\
ISIC2018  & 0.662\(\pm\)0.009 & 0.688\(\pm\)0.014 & 0.687\(\pm\)0.007 & 0.668\(\pm\)0.011 & 0.739\(\pm\)0.018  & 0.797\(\pm\)0.001 & 0.773\(\pm\)0.007  & 0.701\(\pm\)0.003    \\   
\hline
\end{tabular}
}
\end{table*}

\begin{table*}[!t]
\caption{Performance Comparison of Box Prompt on COCO, Fundus, COVID QU-Ex, and ISIC2018 datasets. (Unit: Dice score) }
\centering
\label{tab:SAM_Med_Box}
\resizebox{\linewidth}{!}{
\begin{tabular}{lccccc}
\hline
Dataset   & \begin{tabular}[c]{@{}c@{}}Inner Box of GT mask\\ (Initial Box Result)\end{tabular} & \begin{tabular}[c]{@{}c@{}}Outer Box of  \\ GT mask\end{tabular} & \begin{tabular}[c]{@{}c@{}}Augmented Inner Box from \\ Initial Box Result\end{tabular} & \begin{tabular}[c]{@{}c@{}}Augmented Outer Box from \\ Initial Box Result\end{tabular}& \begin{tabular}[c]{@{}c@{}}Augmented Outer Box from \\ Initial Point Result\end{tabular}\\
\hline
\noalign{\smallskip}
COCO  & 0.106\(\pm\)0.001    &  0.890\(\pm\)0.002   & 0.016\(\pm\)0.001 &0.191\(\pm\)0.002&0.613\(\pm\)0.010\\   
Fundus  & 0.157\(\pm\)0.007    &  0.904\(\pm\)0.002    & 0.013\(\pm\)0.002     &0.244\(\pm\)0.008&0.765\(\pm\)0.014\\
COVID QU-Ex    & 0.110\(\pm\)0.002        &0.744\(\pm\)0.000     & 0.027\(\pm\)0.002       & 0.217\(\pm\)0.008&0.488\(\pm\)0.003\\
ISIC2018  & 0.135\(\pm\)0.008    & 0.883\(\pm\)0.000   & 0.023\(\pm\)0.002 &0.243\(\pm\)0.006&0.653\(\pm\)0.014\\   
\hline
\end{tabular}
}
\end{table*}
\subsection{Comparison of SAM performance on point prompt with and without augmentation}
\label{table_result_analysis}
Table \ref{tab:SAM_Med_Point} presents an exhaustive comparison of the point prompt-based segmentation performance using SAM, with just one point prompt (Column "Initial"), or with augmented point prompts sampled via different strategies as described in Section \ref{aug_random}-\ref{aug_saliency}. Columns "GT Random", "GT Max Entropy", and "GT Max Distance" show the SAM performance by adding the additional point prompt sampled from the ground-truth segmentation mask rather than the initial segmentation mask using different strategies. The performance of these three columns can be regarded as the upper bound of segmentation performance using SAM with two point prompts, as both point prompts are from the ground truth. The "Saliency" sampling strategy does not have its "GT" counterpart, as the saliency map only depends on the input image itself. From the table, it can be found that SAMAug improved the SAM performance on the COCO dataset by 0.01-0.05 as measured by Dice Score without any additional input from the human user. Such performance gain is not by chance, as SAM segmentation using two point prompts sampled from ground truth (listed in the last three columns) can achieve nearly 0.2 performance gain, demonstrating the importance of the additional point prompts. Also, on the COCO dataset, "Max Distance" sampling strategy will lead to the best result. More detailed segmentation performance for the sub-categories in the COCO dataset can be found in Appendix Table I.

For the Fundus dataset, SAMAug can improve the SAM performance by a similar 0.03-0.04 Dice Score, with the "Max Distance" sampling method once again being the best strategy (0.802). For this dataset, an additional point prompt from the ground truth mask has a lower impact than the COCO dataset (around 0.05-0.1 in Dice Score), possibly due to the relatively easier segmentation task involved (just the fundus disc).

For the COVID QU-Ex dataset, the base SAM model achieves a Dice score of 0.488. Compared to this score, SAM segmentation with augmented prompt will improve the performance by around 0.01. It is worth noting that because of the unique segmentation target (separate left and right lung lobes), almost all the methods will achieve a Dice Score of around 0.5, as only one lobe will likely be segmented by SAM.

For the ISIC2018 dataset, the segmentation result by using SAMAug also outperformed the base SAM model by 0.02-0.07. This time, "Saliency" achieved the best performance by a large margin, possibly because the boundary of the target lesion is very irregular, which makes it necessary to sample the point potentially outside the initial result. 

\begin{figure*}[!t]
\centering
\includegraphics[width=0.95\textwidth]{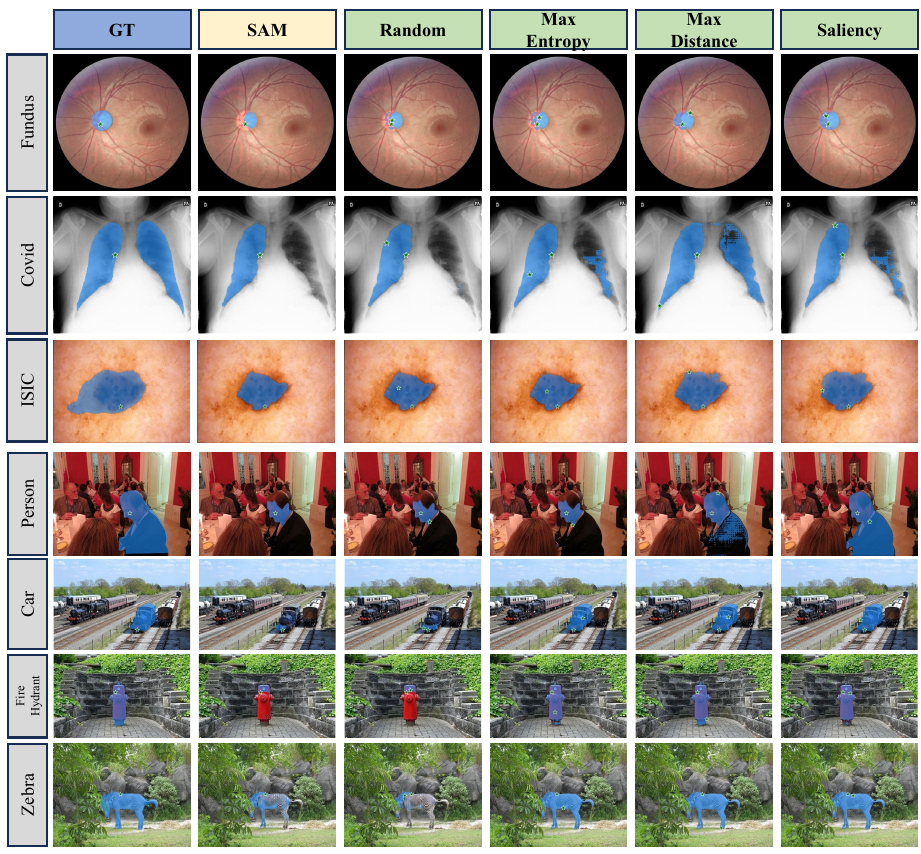}
\caption{Sample segmentation results by SAM with different point prompt augmentation strategies. Column "GT" shows the ground truth segmentation mask. Column "SAM" is the segmentation result using a single point prompt.}
\label{fig:SAM_quality}
\end{figure*}

More visualizations can be found in Figure~\ref{fig:SAM_quality}. Rows one to three show the visualization results of our point prompt augmentation methods on Fundus, COVID QU-Ex, and ISIC2018 datasets. Rows four to seven show the visualization results of four subcategories of the COCO dataset. Each row in the Figure~\ref{fig:SAM_quality} represents the same example. The first column shows the ground-truth mask and the initial point prompt. The second column shows the segmentation result of SAM based on the initial point prompt. Columns three to six show the results of our point prompt augmentation methods (Random, Max Entropy, Max Distance, and Saliency).

\subsection{Comparison of SAM performance on box prompt with and without augmentation}
Table~\ref{tab:SAM_Med_Box} shows the segmentation performance using SAM with box prompts as described in Section~\ref{method}. Column two shows the SAM performance using the box prompt sampled within the GT mask (inner). Columns three to six show the performance using the corresponding augmentation approach. It can be observed that the values in the column ("Outer Box of GT mask") are all the highest, reaching 0.89 on the COCO dataset, 0.904 on the Fundus dataset, 0.744 on the COVID dataset and 0.883 on the ISIC2018 dataset. This result shows that using the outer bounding box (i.e., containing the target region) as a prompt is superior to using a point prompt. However, the prerequisite for using this box is the availability of the full ground-truth masks of the image. In contrast, using the box prompt derived from a point ("Inner Box") will lead to much inferior performance, indicating that SAM is assuming that the provided bounding box is the "Outer Box". We also attempted to generate a second, augmented bounding box prompt from the initial segmentation result, as it is infeasible and meaningless to augment the outer box. It can be found in the third column ("Augmented Inner Box from Initial Box Result") that doing so actually decreases model performance, once again due to the mechanism of how SAM deals with bounding box prompts. Adding the augmented outer box containing the initial result by the inner box prompt will slightly improve the segmentation performance. Yet the result is still much inferior to using point-based prompt augmentation strategies. On the other hand, adding the augmented outer box containing the point prompt-based segmentation result will lead to better performance compared with using the point prompt alone (Table \ref{tab:SAM_Med_Point}, second column "Initial"), albeit still worse than an augmented point prompt (Table \ref{tab:SAM_Med_Point}, fourth to sixth column). In summary, providing a bounding-box prompt will generally lead to better performance of SAM than providing a point-based prompt, although the process involves more interactive actions by the user. On the other hand, prompt augmentation by generating a bounding box containing the initial SAM segmentation result will not work well, even with a good initial result (e.g., segmentation by a point prompt). Detailed performance comparison of COCO sub-categories can be found in Appendix Table II.

\begin{figure}[!t]
\centering
\includegraphics[width=0.95\textwidth]{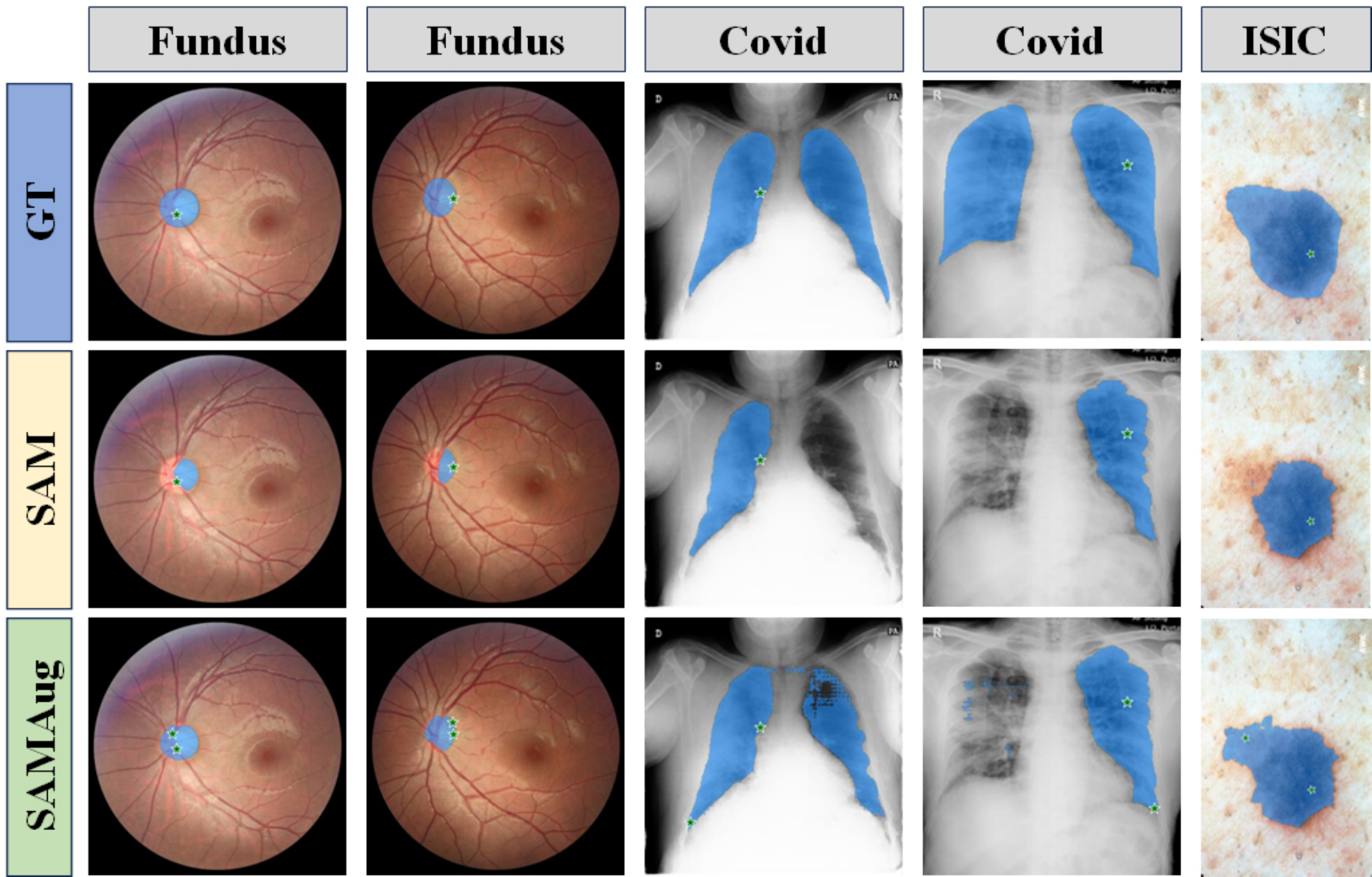}
\caption{Running examples of segmentation results by SAM without (second column) and with (third column) SAMAug. Ground truth segmentation masks are visualized in the first column.}
\label{fig:running_example}
\end{figure}

\subsection{Analysis of Running Examples of SAMAug}

Figure \ref{fig:running_example} provides visual comparative results of SAMAug on three medical datasets. The first two rows belong to the Fundus dataset, the third and fourth rows to COVID QU-Ex, and the fifth and sixth rows to the ISIC dataset. The three columns in the figure correspond to "ground-truth mask," "Result of SAM based on one point," and "Result of SAMAug (SAM with one point prompt and one augmented point prompt)"

In the Fundus dataset, it is evident that SAM based on the initial point prompt often predicts only half of the ground truth. This is primarily due to the distinct edges of vessels, which can easily be misidentified as the boundaries of the segmentation area. However, the situation notably improved after incorporating the additional augmented point prompt. 

For the COVID dataset, the task involves lung region segmentation. As the lung regions are separated into the left and right lobes, most of these segmentation results focus on just one lobe. However, in some cases, the SAM with augmented point prompt will segment both the left and right lung lobes, as demonstrated in the third row of Figure \ref{fig:running_example}. This result demonstrates that SAM learns lung-related features. With the second point prompt at the edge of the lobe, it can then correctly segment both lobes. 

Finally, on the ISIC2018 dataset (last row), it can be observed that the "Saliency" strategy can sample the augmented point prompt beyond the initial result and achieve better performance, which is also indicated by the performance comparison in Table \ref{tab:SAM_Med_Point}. In such cases, SAMAug can effectively mitigate the challenge of complex lesion boundaries and uneven textures of the image, resulting in much-improved segmentation performance.

\section{Discussion and Conclusion}
In this work, we have developed and evaluated SAMAug, a visual point augmentation method designed for SAM. Through this research, we aim to contribute to the ongoing efforts of visual foundation models via a better understanding of the human user's intentions. Based on observations from the experiment, we proposed the concept of invariance in prompts, indicating that while the human user will expect that different prompt selections will lead to similar segmentation results, such invariance needs to be explicitly modeled by augmentation, similar to the premise in traditional data augmentation methods. The improved SAM performance without additional user input, model training/fine-tuning, or adaptation to the data demonstrated the value of visual prompt augmentation for in-context learning-based methods such as SAM and the validity of our prompt invariance assumption. Further experiments on selecting prompts other than the top ones in each sampling strategy, as listed in Appendix Table IV, show that the performance of SAMAug is consistent regarding to the randomness in prompt selection. Overall, SAMAug enriches the human input provided to SAM, enabling a better understanding of the segmentation target by the SAM model and enhancing its performance. 

During our experiments, we found that performance gains from prompt augmentation vary a lot across datasets, indicating the need to determine optimal augmentation approaches for different applications in a more systematic approach. Exploring the use of other metrics or creating hybrid approaches that combine the strengths of multiple techniques. It might also be beneficial to investigate adaptive point augmentation methods that can adjust the strategy based on the specific characteristics of each image or segmentation task. Moreover, the experiment results in Appendix Table III show that adding more than one point prompt using the current scheme will decrease segmentation accuracy. To implement an iterative method for refining the results via multiple rounds of augmentation, we will need to re-design the point prompt sampling strategy, e.g., providing negative samples or point prompts to a different region of interest. 

One promising future direction of SAMAug is its integration with an active learning framework. The interactive nature of SAM makes it suitable for active learning, where we can train a prompt generation model for sampling the most informative point prompts instead of the current rule-based strategies. Furthermore, as SAM can perform segmentation without any user input, it is possible and interesting to investigate whether we can enhance its performance using the active learning-trained prompt sampling model in a fully automatic way. 

The principles and techniques presented in SAMAug have the potential to be beneficial beyond this domain. Large pre-trained models have been successfully applied in various language tasks such as the naming entity recognition \cite{rezayi2022clinicalradiobert,liu2023deid}, relation extraction \cite{cai2022coarse}, question answering \cite{shi2023mededit}, text generation \cite{ma2023impressiongpt}, and text classification \cite{liao2023differentiate,liao2023mask}. Domain-specific LLMs have been developed across various medical applications, including radiology, oncology, clinical trials as well \cite{liu2023tailoring,liu2023radiology,liu2023radiology1,liu2023radonc,holmes2023evaluating,cai2023multimodal,guan2023cohortgpt}. Thus, computer vision large pre-trained models can also be expanded to a wide range of tasks such as object detection, image classification, image synthesis, style transfer, etc., where visual prompt augmentation can be a critical component to these models.  

In addition, an exciting direction for future work of prompt augmentation could be the field of multimodal learning, particularly with large multimodal models that use both image and text prompts \cite{liu2023summary,liu2022survey}. For instance, we could use SAMAug to provide augmented visual prompts while using techniques similar to AugGPT to augment text prompts. The combination of augmented image and text prompts could potentially provide rich information to the model, improving its performance in tasks such as image captioning, visual question answering, or even more complex tasks like multimodal storytelling. The synergistic interaction between visual and textual prompts in such multimodal models could be a fascinating area of exploration in the age of large generative AI models \cite{liu2023summary,zhao2023brain}.

\bibliographystyle{unsrt}  
\bibliography{mybib}

\begin{thebibliography}{10}

\bibitem{liu2023summary}
Yiheng Liu, Tianle Han, Siyuan Ma, Jiayue Zhang, Yuanyuan Yang, Jiaming Tian,
  Hao He, Antong Li, Mengshen He, Zhengliang Liu, et~al.
\newblock Summary of chatgpt/gpt-4 research and perspective towards the future
  of large language models.
\newblock {\em arXiv preprint arXiv:2304.01852}, 2023.

\bibitem{zhao2023brain}
Lin Zhao, Lu~Zhang, Zihao Wu, Yuzhong Chen, Haixing Dai, Xiaowei Yu, Zhengliang
  Liu, Tuo Zhang, Xintao Hu, Xi~Jiang, et~al.
\newblock When brain-inspired ai meets agi.
\newblock {\em arXiv preprint arXiv:2303.15935}, 2023.

\bibitem{willemink2022toward}
Martin~J Willemink, Holger~R Roth, and Veit Sandfort.
\newblock Toward foundational deep learning models for medical imaging in the
  new era of transformer networks.
\newblock {\em Radiology: Artificial Intelligence}, 4(6):e210284, 2022.

\bibitem{zhou2023comprehensive}
Ce~Zhou, Qian Li, Chen Li, Jun Yu, Yixin Liu, Guangjing Wang, Kai Zhang, Cheng
  Ji, Qiben Yan, Lifang He, et~al.
\newblock A comprehensive survey on pretrained foundation models: A history
  from bert to chatgpt.
\newblock {\em arXiv preprint arXiv:2302.09419}, 2023.

\bibitem{kirillov2023segment}
Alexander Kirillov, Eric Mintun, Nikhila Ravi, Hanzi Mao, Chloe Rolland, Laura
  Gustafson, Tete Xiao, Spencer Whitehead, Alexander~C Berg, Wan-Yen Lo, et~al.
\newblock Segment anything.
\newblock {\em arXiv preprint arXiv:2304.02643}, 2023.

\bibitem{yuan2023power}
Yang Yuan.
\newblock On the power of foundation models.
\newblock In {\em International Conference on Machine Learning}, pages
  40519--40530. PMLR, 2023.

\bibitem{zhang2023biomedgpt}
Kai Zhang, Jun Yu, Zhiling Yan, Yixin Liu, Eashan Adhikarla, Sunyang Fu, Xun
  Chen, Chen Chen, Yuyin Zhou, Xiang Li, et~al.
\newblock Biomedgpt: A unified and generalist biomedical generative pre-trained
  transformer for vision, language, and multimodal tasks.
\newblock {\em arXiv preprint arXiv:2305.17100}, 2023.

\bibitem{brown2020language}
Tom Brown, Benjamin Mann, Nick Ryder, Melanie Subbiah, Jared~D Kaplan, Prafulla
  Dhariwal, Arvind Neelakantan, Pranav Shyam, Girish Sastry, Amanda Askell,
  et~al.
\newblock Language models are few-shot learners.
\newblock {\em Advances in neural information processing systems},
  33:1877--1901, 2020.

\bibitem{zhang2023comprehensive}
Chunhui Zhang, Li~Liu, Yawen Cui, Guanjie Huang, Weilin Lin, Yiqian Yang, and
  Yuehong Hu.
\newblock A comprehensive survey on segment anything model for vision and
  beyond.
\newblock {\em arXiv preprint arXiv:2305.08196}, 2023.

\bibitem{li2023artificial}
Xiang Li, Lu~Zhang, Zihao Wu, Zhengliang Liu, Lin Zhao, Yixuan Yuan, Jun Liu,
  Gang Li, Dajiang Zhu, Pingkuan Yan, et~al.
\newblock Artificial general intelligence for medical imaging.
\newblock {\em arXiv preprint arXiv:2306.05480}, 2023.

\bibitem{zhang2023segment}
Lian Zhang, Zhengliang Liu, Lu~Zhang, Zihao Wu, Xiaowei Yu, Jason Holmes,
  Hongying Feng, Haixing Dai, Xiang Li, Quanzheng Li, et~al.
\newblock Segment anything model (sam) for radiation oncology.
\newblock {\em arXiv preprint arXiv:2306.11730}, 2023.

\bibitem{deng2023segment}
Ruining Deng, Can Cui, Quan Liu, Tianyuan Yao, Lucas~W Remedios, Shunxing Bao,
  Bennett~A Landman, Lee~E Wheless, Lori~A Coburn, Keith~T Wilson, et~al.
\newblock Segment anything model (sam) for digital pathology: Assess zero-shot
  segmentation on whole slide imaging.
\newblock {\em arXiv preprint arXiv:2304.04155}, 2023.

\bibitem{he2023accuracy}
Sheng He, Rina Bao, Jingpeng Li, P~Ellen Grant, and Yangming Ou.
\newblock Accuracy of segment-anything model (sam) in medical image
  segmentation tasks.
\newblock {\em arXiv preprint arXiv:2304.09324}, 2023.

\bibitem{zhang2023input}
Yizhe Zhang, Tao Zhou, Peixian Liang, and Danny~Z Chen.
\newblock Input augmentation with sam: Boosting medical image segmentation with
  segmentation foundation model.
\newblock {\em arXiv preprint arXiv:2304.11332}, 2023.

\bibitem{wu2023medical}
Junde Wu, Rao Fu, Huihui Fang, Yuanpei Liu, Zhaowei Wang, Yanwu Xu, Yueming
  Jin, and Tal Arbel.
\newblock Medical sam adapter: Adapting segment anything model for medical
  image segmentation.
\newblock {\em arXiv preprint arXiv:2304.12620}, 2023.

\bibitem{chen2023ma}
Cheng Chen, Juzheng Miao, Dufan Wu, Zhiling Yan, Sekeun Kim, Jiang Hu, Aoxiao
  Zhong, Zhengliang Liu, Lichao Sun, Xiang Li, et~al.
\newblock Ma-sam: Modality-agnostic sam adaptation for 3d medical image
  segmentation.
\newblock {\em arXiv preprint arXiv:2309.08842}, 2023.

\bibitem{kim2023medivista}
Sekeun Kim, Kyungsang Kim, Jiang Hu, Cheng Chen, Zhiliang Lyu, Ren Hui,
  Sunghwan Kim, Zhengliang Liu, Aoxiao Zhong, Xiang Li, et~al.
\newblock Medivista-sam: Zero-shot medical video analysis with spatio-temporal
  sam adaptation.
\newblock {\em arXiv preprint arXiv:2309.13539}, 2023.

\bibitem{huang2023segment}
Yuhao Huang, Xin Yang, Lian Liu, Han Zhou, Ao~Chang, Xinrui Zhou, Rusi Chen,
  Junxuan Yu, Jiongquan Chen, Chaoyu Chen, et~al.
\newblock Segment anything model for medical images?
\newblock {\em arXiv preprint arXiv:2304.14660}, 2023.

\bibitem{diao2022black}
Shizhe Diao, Zhichao Huang, Ruijia Xu, Xuechun Li, Yong Lin, Xiao Zhou, and
  Tong Zhang.
\newblock Black-box prompt learning for pre-trained language models.
\newblock {\em arXiv preprint arXiv:2201.08531}, 2022.

\bibitem{wang2023prompt}
Jiaqi Wang, Enze Shi, Sigang Yu, Zihao Wu, Chong Ma, Haixing Dai, Qiushi Yang,
  Yanqing Kang, Jinru Wu, Huawen Hu, et~al.
\newblock Prompt engineering for healthcare: Methodologies and applications.
\newblock {\em arXiv preprint arXiv:2304.14670}, 2023.

\bibitem{gao2020making}
Tianyu Gao, Adam Fisch, and Danqi Chen.
\newblock Making pre-trained language models better few-shot learners.
\newblock {\em arXiv preprint arXiv:2012.15723}, 2020.

\bibitem{ben2022pada}
Eyal Ben-David, Nadav Oved, and Roi Reichart.
\newblock Pada: Example-based prompt learning for on-the-fly adaptation to
  unseen domains.
\newblock {\em Transactions of the Association for Computational Linguistics},
  10:414--433, 2022.

\bibitem{qin2021learning}
Guanghui Qin and Jason Eisner.
\newblock Learning how to ask: Querying lms with mixtures of soft prompts.
\newblock {\em arXiv preprint arXiv:2104.06599}, 2021.

\bibitem{liu2021gpt}
Xiao Liu, Yanan Zheng, Zhengxiao Du, Ming Ding, Yujie Qian, Zhilin Yang, and
  Jie Tang.
\newblock Gpt understands, too.
\newblock {\em arXiv preprint arXiv:2103.10385}, 2021.

\bibitem{shum2023automatic}
KaShun Shum, Shizhe Diao, and Tong Zhang.
\newblock Automatic prompt augmentation and selection with chain-of-thought
  from labeled data.
\newblock {\em arXiv preprint arXiv:2302.12822}, 2023.

\bibitem{dai2023chataug}
Haixing Dai, Zhengliang Liu, Wenxiong Liao, Xiaoke Huang, Zihao Wu, Lin Zhao,
  Wei Liu, Ninghao Liu, Sheng Li, Dajiang Zhu, et~al.
\newblock Chataug: Leveraging chatgpt for text data augmentation.
\newblock {\em arXiv preprint arXiv:2302.13007}, 2023.

\bibitem{mcguinness2010comparative}
Kevin McGuinness and Noel~E O’connor.
\newblock A comparative evaluation of interactive segmentation algorithms.
\newblock {\em Pattern Recognition}, 43(2):434--444, 2010.

\bibitem{li2015interactive}
Xiang Li, Zhi Zhou, Philipp Keller, Hongkui Zeng, Tianming Liu, and Hanchuan
  Peng.
\newblock Interactive exemplar-based segmentation toolkit for biomedical image
  analysis.
\newblock In {\em 2015 IEEE 12th International Symposium on Biomedical Imaging
  (ISBI)}, pages 168--171. IEEE, 2015.

\bibitem{xiao2023instruction}
Zhenxiang Xiao, Yuzhong Chen, Lu~Zhang, Junjie Yao, Zihao Wu, Xiaowei Yu,
  Yi~Pan, Lin Zhao, Chong Ma, Xinyu Liu, et~al.
\newblock Instruction-vit: Multi-modal prompts for instruction learning in vit.
\newblock {\em arXiv preprint arXiv:2305.00201}, 2023.

\bibitem{jia2022visual}
Menglin Jia, Luming Tang, Bor-Chun Chen, Claire Cardie, Serge Belongie, Bharath
  Hariharan, and Ser-Nam Lim.
\newblock Visual prompt tuning.
\newblock In {\em European Conference on Computer Vision}, pages 709--727.
  Springer, 2022.

\bibitem{jie2022convolutional}
Shibo Jie and Zhi-Hong Deng.
\newblock Convolutional bypasses are better vision transformer adapters.
\newblock {\em arXiv preprint arXiv:2207.07039}, 2022.

\bibitem{zhu2023visual}
Jiawen Zhu, Simiao Lai, Xin Chen, Dong Wang, and Huchuan Lu.
\newblock Visual prompt multi-modal tracking.
\newblock In {\em Proceedings of the IEEE/CVF Conference on Computer Vision and
  Pattern Recognition}, pages 9516--9526, 2023.

\bibitem{berndt2020sampling}
Andrea~E Berndt.
\newblock Sampling methods.
\newblock {\em Journal of Human Lactation}, 36(2):224--226, 2020.

\bibitem{wu2017sampling}
Chao-Yuan Wu, R~Manmatha, Alexander~J Smola, and Philipp Krahenbuhl.
\newblock Sampling matters in deep embedding learning.
\newblock In {\em Proceedings of the IEEE international conference on computer
  vision}, pages 2840--2848, 2017.

\bibitem{lowe2004distinctive}
David~G Lowe.
\newblock Distinctive image features from scale-invariant keypoints.
\newblock {\em International journal of computer vision}, 60:91--110, 2004.

\bibitem{bay2006surf}
Herbert Bay, Tinne Tuytelaars, and Luc Van~Gool.
\newblock Surf: Speeded up robust features.
\newblock {\em Lecture notes in computer science}, 3951:404--417, 2006.

\bibitem{ma2019lmr}
Jiayi Ma, Xingyu Jiang, Junjun Jiang, Ji~Zhao, and Xiaojie Guo.
\newblock Lmr: Learning a two-class classifier for mismatch removal.
\newblock {\em IEEE Transactions on Image Processing}, 28(8):4045--4059, 2019.

\bibitem{ozden2005image}
Mustafa {\"O}zden and Ediz Polat.
\newblock Image segmentation using color and texture features.
\newblock In {\em 2005 13th European Signal Processing Conference}, pages 1--4.
  IEEE, 2005.

\bibitem{zhao2015image}
Chunlan Zhao.
\newblock Image segmentation based on fast normalized cut.
\newblock {\em The Open Cybernetics \& Systemics Journal}, 9(1), 2015.

\bibitem{long2015fully}
Jonathan Long, Evan Shelhamer, and Trevor Darrell.
\newblock Fully convolutional networks for semantic segmentation.
\newblock In {\em Proceedings of the IEEE conference on computer vision and
  pattern recognition}, pages 3431--3440, 2015.

\bibitem{itti1998model}
Laurent Itti, Christof Koch, and Ernst Niebur.
\newblock A model of saliency-based visual attention for rapid scene analysis.
\newblock {\em IEEE Transactions on pattern analysis and machine intelligence},
  20(11):1254--1259, 1998.

\bibitem{achanta2008salient}
Radhakrishna Achanta, Francisco Estrada, Patricia Wils, and Sabine
  S{\"u}sstrunk.
\newblock Salient region detection and segmentation.
\newblock In {\em Computer Vision Systems: 6th International Conference, ICVS
  2008 Santorini, Greece, May 12-15, 2008 Proceedings 6}, pages 66--75.
  Springer, 2008.

\bibitem{tong2015salient}
Na~Tong, Huchuan Lu, Ying Zhang, and Xiang Ruan.
\newblock Salient object detection via global and local cues.
\newblock {\em Pattern Recognition}, 48(10):3258--3267, 2015.

\bibitem{ma2023rectify}
Chong Ma, Lin Zhao, Yuzhong Chen, Lei Guo, Tuo Zhang, Xintao Hu, Dinggang Shen,
  Xi~Jiang, and Tianming Liu.
\newblock Rectify vit shortcut learning by visual saliency.
\newblock {\em IEEE Transactions on Neural Networks and Learning Systems},
  2023.

\bibitem{alvarez1993axioms}
Luis Alvarez, Fr{\'e}d{\'e}ric Guichard, Pierre~Louis Lions, and Jean~Michel
  Morel.
\newblock Axioms and fundamental equations of image processing.
\newblock {\em Archive for rational mechanics and analysis}, 123:199--257,
  1993.

\bibitem{liu2021visual}
Nian Liu, Ni~Zhang, Kaiyuan Wan, Ling Shao, and Junwei Han.
\newblock Visual saliency transformer.
\newblock In {\em Proceedings of the IEEE/CVF international conference on
  computer vision}, pages 4722--4732, 2021.

\bibitem{cocodataset}
Tsung{-}Yi Lin, Michael Maire, Serge~J. Belongie, Lubomir~D. Bourdev, Ross~B.
  Girshick, James Hays, Pietro Perona, Deva Ramanan, Piotr Doll{'{a} }r, and
  C.~Lawrence Zitnick.
\newblock Microsoft {COCO:} common objects in context.
\newblock {\em CoRR}, abs/1405.0312, 2014.

\bibitem{eyepacs_2012}
EyePACS.
\newblock Eyepacs dataset.
\newblock Data set, 2012.
\newblock Kaggle.

\bibitem{codella2019skin}
Noel Codella, Veronica Rotemberg, Philipp Tschandl, M~Emre Celebi, Stephen
  Dusza, David Gutman, Brian Helba, Aadi Kalloo, Konstantinos Liopyris, Michael
  Marchetti, et~al.
\newblock Skin lesion analysis toward melanoma detection 2018: A challenge
  hosted by the international skin imaging collaboration (isic).
\newblock {\em arXiv preprint arXiv:1902.03368}, 2019.

\bibitem{tahir3122958covid}
Anas~M Tahir, Muhammad~EH Chowdhury, Yazan Qiblawey, Amith Khandakar, Tawsifur
  Rahman, Serkan Kiranyaz, Uzair Khurshid, Nabil Ibtehaz, Sakib Mahmud, and
  Maymouna Ezeddin.
\newblock Covid-qu-ex dataset, 2022.
\newblock {\em URL https://www. kaggle. com/dsv/3122958}, 11, 2022.

\bibitem{rezayi2022clinicalradiobert}
Saed Rezayi, Haixing Dai, Zhengliang Liu, Zihao Wu, Akarsh Hebbar, Andrew~H
  Burns, Lin Zhao, Dajiang Zhu, Quanzheng Li, Wei Liu, et~al.
\newblock Clinicalradiobert: Knowledge-infused few shot learning for clinical
  notes named entity recognition.
\newblock In {\em International Workshop on Machine Learning in Medical
  Imaging}, pages 269--278. Springer, 2022.

\bibitem{liu2023deid}
Zhengliang Liu, Xiaowei Yu, Lu~Zhang, Zihao Wu, Chao Cao, Haixing Dai, Lin
  Zhao, Wei Liu, Dinggang Shen, Quanzheng Li, et~al.
\newblock Deid-gpt: Zero-shot medical text de-identification by gpt-4.
\newblock {\em arXiv preprint arXiv:2303.11032}, 2023.

\bibitem{cai2022coarse}
Homgmin Cai, Wenxiong Liao, Zhengliang Liu, Xiaoke Huang, Yiyang Zhang, Siqi
  Ding, Sheng Li, Quanzheng Li, Tianming Liu, and Xiang Li.
\newblock Coarse-to-fine knowledge graph domain adaptation based on
  distantly-supervised iterative training.
\newblock {\em arXiv preprint arXiv:2211.02849}, 2022.

\bibitem{shi2023mededit}
Yucheng Shi, Shaochen Xu, Zhengliang Liu, Tianming Liu, Xiang Li, and Ninghao
  Liu.
\newblock Mededit: Model editing for medical question answering with external
  knowledge bases.
\newblock {\em arXiv preprint arXiv:2309.16035}, 2023.

\bibitem{ma2023impressiongpt}
Chong Ma, Zihao Wu, Jiaqi Wang, Shaochen Xu, Yaonai Wei, Zhengliang Liu, Lei
  Guo, Xiaoyan Cai, Shu Zhang, Tuo Zhang, et~al.
\newblock Impressiongpt: an iterative optimizing framework for radiology report
  summarization with chatgpt.
\newblock {\em arXiv preprint arXiv:2304.08448}, 2023.

\bibitem{liao2023differentiate}
Wenxiong Liao, Zhengliang Liu, Haixing Dai, Shaochen Xu, Zihao Wu, Yiyang
  Zhang, Xiaoke Huang, Dajiang Zhu, Hongmin Cai, Tianming Liu, et~al.
\newblock Differentiate chatgpt-generated and human-written medical texts.
\newblock {\em arXiv preprint arXiv:2304.11567}, 2023.

\bibitem{liao2023mask}
Wenxiong Liao, Zhengliang Liu, Haixing Dai, Zihao Wu, Yiyang Zhang, Xiaoke
  Huang, Yuzhong Chen, Xi~Jiang, Dajiang Zhu, Tianming Liu, et~al.
\newblock Mask-guided bert for few shot text classification.
\newblock {\em arXiv preprint arXiv:2302.10447}, 2023.

\bibitem{liu2023tailoring}
Zhengliang Liu, Aoxiao Zhong, Yiwei Li, Longtao Yang, Chao Ju, Zihao Wu, Chong
  Ma, Peng Shu, Cheng Chen, Sekeun Kim, et~al.
\newblock Tailoring large language models to radiology: A preliminary approach
  to llm adaptation for a highly specialized domain.
\newblock In {\em International Workshop on Machine Learning in Medical
  Imaging}, pages 464--473. Springer, 2023.

\bibitem{liu2023radiology}
Zhengliang Liu, Aoxiao Zhong, Yiwei Li, Longtao Yang, Chao Ju, Zihao Wu, Chong
  Ma, Peng Shu, Cheng Chen, Sekeun Kim, et~al.
\newblock Radiology-gpt: A large language model for radiology.
\newblock {\em arXiv preprint arXiv:2306.08666}, 2023.

\bibitem{liu2023radiology1}
Zhengliang Liu, Yiwei Li, Peng Shu, Aoxiao Zhong, Longtao Yang, Chao Ju, Zihao
  Wu, Chong Ma, Jie Luo, Cheng Chen, et~al.
\newblock Radiology-llama2: Best-in-class large language model for radiology.
\newblock {\em arXiv preprint arXiv:2309.06419}, 2023.

\bibitem{liu2023radonc}
Zhengliang Liu, Peilong Wang, Yiwei Li, Jason Holmes, Peng Shu, Lian Zhang,
  Chenbin Liu, Ninghao Liu, Dajiang Zhu, Xiang Li, et~al.
\newblock Radonc-gpt: A large language model for radiation oncology.
\newblock {\em arXiv preprint arXiv:2309.10160}, 2023.

\bibitem{holmes2023evaluating}
Jason Holmes, Zhengliang Liu, Lian Zhang, Yuzhen Ding, Terence~T Sio, Lisa~A
  McGee, Jonathan~B Ashman, Xiang Li, Tianming Liu, Jiajian Shen, et~al.
\newblock Evaluating large language models on a highly-specialized topic,
  radiation oncology physics.
\newblock {\em arXiv preprint arXiv:2304.01938}, 2023.

\bibitem{cai2023multimodal}
Hongmin Cai, Xiaoke Huang, Zhengliang Liu, Wenxiong Liao, Haixing Dai, Zihao
  Wu, Dajiang Zhu, Hui Ren, Quanzheng Li, Tianming Liu, et~al.
\newblock Multimodal approaches for alzheimer’s detection using patients’
  speech and transcript.
\newblock In {\em International Conference on Brain Informatics}, pages
  395--406. Springer, 2023.

\bibitem{guan2023cohortgpt}
Zihan Guan, Zihao Wu, Zhengliang Liu, Dufan Wu, Hui Ren, Quanzheng Li, Xiang
  Li, and Ninghao Liu.
\newblock Cohortgpt: An enhanced gpt for participant recruitment in clinical
  study.
\newblock {\em arXiv preprint arXiv:2307.11346}, 2023.

\bibitem{liu2022survey}
Zhengliang Liu, Mengshen He, Zuowei Jiang, Zihao Wu, Haixing Dai, Lian Zhang,
  Siyi Luo, Tianle Han, Xiang Li, Xi~Jiang, et~al.
\newblock Survey on natural language processing in medical image analysis.
\newblock {\em Zhong nan da xue xue bao. Yi xue ban= Journal of Central South
  University. Medical Sciences}, 47(8):981--993, 2022.

\end{thebibliography}

\begin{appendices}

\setcounter{table}{0}
\section{Comparison on Point Prompt}
The comparison of model performance with/without point prompts on subcategories of the COCO dataset is shown in Table \ref{tab:appendix_SAM_Med_Point}. In the "person" category, the base SAM model yields a Dice score of 0.4552. With the "Random" augmentation method, the performance improved to 0.4594, while Max Entropy achieved a more noticeable lift to 0.4677. The Max Distance method significantly boosts the Dice Score to 0.51, and the Saliency method further extends the improvement, achieving the highest score of 0.5535. Similar performance improvement trends can be observed across all other subcategories of the COCO dataset. For instance, in the "COCO" of Table \ref{tab:appendix_SAM_Med_Point}, the base SAM model's Dice (Initial) score of 0.601 is exceeded by all our methods, with the Max Distance method achieving the highest score of 0.651.

\section{Comparison on Box Prompt}
Table~\ref{tab:appendix_SAM_Med_Box} shows the results using box prompt. Specifically, columns two to six show the results of box prompt. The top half of Table \ref{tab:appendix_SAM_Med_Box} shows the average comparison results on COCO, Fundus, ISIC2018, and Covid CXR datasets. The bottom half of Table \ref{tab:appendix_SAM_Med_Box} shows the results of subcategories in the COCO dataset. 

\renewcommand{\thetable}{\Roman{table}}
\begin{table*}[!t]
\caption{Performance Comparison of Point Augmentations on COCO dataset, different sub-categories. (Unit: Dice score) }
\centering
\label{tab:appendix_SAM_Med_Point}
\resizebox{\linewidth}{!}{
\begin{tabular}{lcccccccc}
\hline
Dataset   & Initial       & Random        & Max Entropy   & Max Distance  & Saliency  & GT Random     & GT Max Entropy & GT Max Distance \\
\hline
\noalign{\smallskip}
person &0.4552\(\pm\)0.0346 & 0.4594\(\pm\)0.0351 & 0.4677\(\pm\)0.0347   & 0.5100\(\pm\)0.0327     & 0.5535\(\pm\)0.0434 & 0.7372\(\pm\)0.0375 & 0.7514\(\pm\)0.0242 & 0.8061\(\pm\)0.0285  \\
skateboard & 0.5770\(\pm\)0.0882   & 0.5805\(\pm\)0.0837    & 0.5979\(\pm\)0.0916     & 0.6238\(\pm\)0.0745     & 0.6964\(\pm\)0.0387 & 0.8268\(\pm\)0.0406 & 0.8174\(\pm\)0.0350 & 0.7310\(\pm\)0.0073  \\
bottle & 0.5722\(\pm\)0.0381         & 0.6026\(\pm\)0.0513     & 0.6135\(\pm\)0.0332   & 0.6512\(\pm\)0.0532       & 0.6474\(\pm\)0.0521 & 0.8257\(\pm\)0.0160 & 0.7783\(\pm\)0.0194 & 0.8517\(\pm\)0.0160 \\
tv  & 0.6409\(\pm\)0.0302     & 0.6437\(\pm\)0.0344     & 0.6832\(\pm\)0.0127       & 0.7530\(\pm\)0.0108       & 0.7266\(\pm\)0.0264  & 0.8421\(\pm\)0.0217 & 0.8665\(\pm\)0.0262 & 0.8682\(\pm\)0.0140 \\
vase  & 0.6533\(\pm\)0.0563     & 0.6680\(\pm\)0.0444       & 0.6949\(\pm\)0.0330    & 0.7079\(\pm\)0.0156       & 0.7287\(\pm\)0.0228 & 0.8797\(\pm\)0.0230 & 0.8401\(\pm\)0.0502 & 0.8641\(\pm\)0.0586 \\
stop sign & 0.7886\(\pm\)0.0135 & 0.7927\(\pm\)0.0201 & 0.7991\(\pm\)0.0195 & 0.8081\(\pm\)0.0213 & 0.8563\(\pm\)0.0247 & 0.9392\(\pm\)0.0490 & 0.9483\(\pm\)0.0311 & 0.9528\(\pm\)0.0174 \\
cat & 0.7580\(\pm\)0.0576 & 0.7740\(\pm\)0.0544 & 0.7797\(\pm\)0.0624 & 0.7734\(\pm\)0.0619 & 0.7584\(\pm\)0.0646 & 0.8990\(\pm\)0.0096 & 0.8481\(\pm\)0.0434 & 0.7948\(\pm\)0.0313 \\
horse & 0.6299\(\pm\)0.0185 & 0.6506\(\pm\)0.0136 & 0.6472\(\pm\)0.0319 & 0.6738\(\pm\)0.0451 & 0.5985\(\pm\)0.0446 & 0.8154\(\pm\)0.0174 & 0.8157\(\pm\)0.0103 & 0.8126\(\pm\)0.0186 \\
umbrella & 0.7496\(\pm\)0.0426 & 0.7593\(\pm\)0.0477 & 0.7768\(\pm\)0.0504 & 0.7922\(\pm\)0.0624 & 0.7800\(\pm\)0.0485 & 0.8715\(\pm\)0.0018 & 0.8764\(\pm\)0.0047 & 0.8521\(\pm\)0.0124 \\
sports ball & 0.6555\(\pm\)0.1222 & 0.6948\(\pm\)0.0853 & 0.6779\(\pm\)0.1126 & 0.6939\(\pm\)0.0726 & 0.7155\(\pm\)0.1129& 0.8150\(\pm\)0.1049 & 0.8170\(\pm\)0.0695 & 0.8224\(\pm\)0.0646 \\
wine glass & 0.5938\(\pm\)0.0324 & 0.6038\(\pm\)0.0439 & 0.6034\(\pm\)0.0511 & 0.6829\(\pm\)0.0499 & 0.6520\(\pm\)0.0428 & 0.7755\(\pm\)0.0293 & 0.7331\(\pm\)0.0246 & 0.8204\(\pm\)0.0086  \\
hot dog & 0.4634\(\pm\)0.0232 & 0.4805\(\pm\)0.0166 & 0.4850\(\pm\)0.0175 & 0.5094\(\pm\)0.0198 & 0.5877\(\pm\)0.0300 & 0.7424\(\pm\)0.0331 & 0.6646\(\pm\)0.0406 & 0.6260\(\pm\)0.0244 \\
chair & 0.5719\(\pm\)0.0156 & 0.5920\(\pm\)0.0099 & 0.6032\(\pm\)0.0195 & 0.6234\(\pm\)0.0155 & 0.6137\(\pm\)0.0317 & 0.7580\(\pm\)0.0516 & 0.7096\(\pm\)0.0369 & 0.7487\(\pm\)0.0115  \\
toilet & 0.5895\(\pm\)0.0552 & 0.6167\(\pm\)0.0473 & 0.6528\(\pm\)0.0516 & 0.6785\(\pm\)0.0607 & 0.6512\(\pm\)0.0585 & 0.8246\(\pm\)0.0079 & 0.8281\(\pm\)0.0232 & 0.7962\(\pm\)0.0582  \\
\noalign{\smallskip}
\hline
\end{tabular}
}
\end{table*}

\begin{table*}[!t]
\caption{Performance Comparison of Box Augmentations on COCO dataset, different sub-categories. (Unit: Dice score) }
\centering
\label{tab:appendix_SAM_Med_Box}
\resizebox{\linewidth}{!}{
\begin{tabular}{lccccc}
\hline
Dataset   & \begin{tabular}[c]{@{}c@{}}Inner Box in GT mask\\ (Initial Box Result)\end{tabular} & \begin{tabular}[c]{@{}c@{}}Outer Box of \\ Initial Box Result\end{tabular} & \begin{tabular}[c]{@{}c@{}}Inner Box in \\ Initial Box Result\end{tabular} & \begin{tabular}[c]{@{}c@{}}Outer Box of  \\ GT mask\end{tabular} & \begin{tabular}[c]{@{}c@{}}Outer Box of  \\ Initial Point Result\end{tabular} \\
\hline
Person & 0.0715\(\pm\)0.0226 & 0.1377\(\pm\)0.0411 & 0.0081\(\pm\)0.0027 & 0.8792\(\pm\)0.0257 & 0.4899\(\pm\)0.0637 \\
skateboard & 0.0954\(\pm\)0.0284 & 0.1630\(\pm\)0.0355 & 0.0108\(\pm\)0.0052 & 0.9025\(\pm\)0.0231 & 0.6113\(\pm\)0.1120 \\
bottle  & 0.1026\(\pm\)0.0073 & 0.2016\(\pm\)0.0295 & 0.0153\(\pm\)0.0016 & 0.8965\(\pm\)0.0266 & 0.6649\(\pm\)0.0454 \\
tv  & 0.1401\(\pm\)0.0291 & 0.2865\(\pm\)0.0155 & 0.0164\(\pm\)0.0048 & 0.9445\(\pm\)0.0125 & 0.7004\(\pm\)0.1017 \\
vase  & 0.1194\(\pm\)0.0289 & 0.2007\(\pm\)0.0606 & 0.0114\(\pm\)0.0032 & 0.9407\(\pm\)0.0037 & 0.7720\(\pm\)0.0302 \\
stop sign  & 0.0984\(\pm\)0.0426 & 0.1735\(\pm\)0.0666 & 0.0125\(\pm\)0.0076 & 0.9753\(\pm\)0.0008 & 0.7949\(\pm\)0.0122 \\
cat  & 0.0877\(\pm\)0.0123 & 0.1831\(\pm\)0.0235 & 0.0066\(\pm\)0.0011 & 0.9361\(\pm\)0.0079 & 0.7642\(\pm\)0.0554 \\
horse  & 0.0617\(\pm\)0.0277 & 0.1250\(\pm\)0.0461 & 0.0057\(\pm\)0.0037 & 0.8988\(\pm\)0.0027 & 0.6639\(\pm\)0.0128 \\
umbrella  & 0.0917\(\pm\)0.0061 & 0.1810\(\pm\)0.0282 & 0.0073\(\pm\)0.0004 & 0.9183\(\pm\)0.0069 & 0.7619\(\pm\)0.0688 \\
sports ball  & 0.0822\(\pm\)0.0338 & 0.1408\(\pm\)0.0289 & 0.0125\(\pm\)0.0130 & 0.9083\(\pm\)0.0456 & 0.6671\(\pm\)0.1055 \\
wine glass & 0.1104\(\pm\)0.0078 & 0.1948\(\pm\)0.0193 & 0.0150\(\pm\)0.0009 & 0.8847\(\pm\)0.0200 & 0.6303\(\pm\)0.0502 \\
hot dog  & 0.0840\(\pm\)0.0139 & 0.1521\(\pm\)0.0196 & 0.0086\(\pm\)0.0032 & 0.8132\(\pm\)0.0412 & 0.5019\(\pm\)0.0263 \\
chair & 0.1161\(\pm\)0.0293 & 0.1643\(\pm\)0.0176 & 0.0202\(\pm\)0.0036 & 0.8388\(\pm\)0.0082 & 0.6044\(\pm\)0.0151 \\
toilet & 0.1160\(\pm\)0.0283 & 0.1961\(\pm\)0.0159 & 0.0222\(\pm\)0.0156 & 0.9365\(\pm\)0.0146 & 0.6568\(\pm\)0.0763 \\
\noalign{\smallskip}
\hline
\end{tabular}
}
\end{table*}

\begin{table*}[!t]
\caption{Performance Comparison of Multiple-point Prompt Augmentation on COCO, Fundus, COVID QU-Ex and ISIC2018 datasets. (Unit: Dice score) }
\centering
\label{tab:appendix_SAM_Point_Num}
\begin{tabular}{lcccccc}
\hline
Dataset   & \multicolumn{1}{c}{Points Number}    & \multicolumn{1}{c}{Initial SAM} & \multicolumn{1}{c}{Random} & \multicolumn{1}{c}{Max Entropy} & \multicolumn{1}{c}{Max Distance} & \multicolumn{1}{c}{Saliency} \\ 
\hline
\noalign{\smallskip}
COCO & 2 points        & 0.6005\(\pm\)0.0019     & 0.6137\(\pm\)0.0024       & 0.6212\(\pm\)0.0039            & 0.6514\(\pm\)0.0021          & 0.6314\(\pm\)0.0008          \\
COCO & 3 points  & 0.5852\(\pm\)0.0022    & 0.6061\(\pm\)0.0011  & 0.6160\(\pm\)0.0007     & 0.6473\(\pm\)0.0011    & 0.6434\(\pm\)0.0064    \\
COCO & 5 points       & 0.5864\(\pm\)0.0043  & 0.6069\(\pm\)0.0040 & 0.6061\(\pm\)0.0046                  & 0.6501\(\pm\)0.0059       & 0.6434\(\pm\)0.0048            \\
Fundus & 2 points     & 0.7662\(\pm\)0.0084       & 0.7939\(\pm\)0.0069       & 0.7909\(\pm\)0.0064      & 0.8022\(\pm\)0.0074       & 0.7921\(\pm\)0.0016         \\
Fundus & 3 points  & 0.7508\(\pm\)0.0071    & 0.7717\(\pm\)0.0051  & 0.7672\(\pm\)0.0011     & 0.7895\(\pm\)0.0038    & 0.7745\(\pm\)0.0027    \\
Fundus & 5 points       & 0.7527\(\pm\)0.0046  & 0.7566\(\pm\)0.0047 & 0.7185\(\pm\)0.0028    & 0.7814\(\pm\)0.0034   & 0.7726\(\pm\)0.0069    \\
COVID QU-Ex  & 2 points   & 0.4879\(\pm\)0.0033 & 0.5032\(\pm\)0.0033 & 0.4900\(\pm\)0.0033 & 0.4968\(\pm\)0.0019 & 0.4949\(\pm\)0.0019      \\
COVID QU-Ex  & 3 points  & 0.4864\(\pm\)0.0030 & 0.4996\(\pm\)0.0032 & 0.4898\(\pm\)0.0032 & 0.4929\(\pm\)0.0039 & 0.4864\(\pm\)0.0012   \\
COVID QU-Ex  & 5 points  & 0.4860\(\pm\)0.0047 & 0.4970\(\pm\)0.0030 & 0.4877\(\pm\)0.0046 & 0.4916\(\pm\)0.0041 & 0.4764\(\pm\)0.0032   \\
ISIC2018  & 2 points   & 0.6620\(\pm\)0.0094 & 0.6880\(\pm\)0.0138 & 0.6869\(\pm\)0.0069 & 0.6679\(\pm\)0.0111 & 0.7386\(\pm\)0.0175   \\
ISIC2018  & 3 points  & 0.6631\(\pm\)0.0029 & 0.6779\(\pm\)0.0045 & 0.6733\(\pm\)0.0019 & 0.6660\(\pm\)0.0032  & 0.7400\(\pm\)0.0057  \\
ISIC2018  & 5 points   & 0.6643\(\pm\)0.0049 & 0.6705\(\pm\)0.0045 & 0.6720\(\pm\)0.0059 & 0.6670\(\pm\)0.0079 & 0.7655\(\pm\)0.0081   \\
\noalign{\smallskip}
\hline
\end{tabular}
\end{table*}

\begin{table*}[t]
\caption{Stability Study of the Point Prompt Selection on Model Performance. (Unit: Dice score) }
\centering
\label{tab:appendix_SAM_Point_Stability}
\begin{tabular}{lcccc}
\hline
\noalign{\smallskip}
Dataset   & \multicolumn{1}{c}{Points Sequence}   & \multicolumn{1}{c}{Max Entropy} & \multicolumn{1}{c}{Max Distance} & \multicolumn{1}{c}{Saliency} \\ \hline
COCO & 1$st$    & 0.6043\(\pm\)0.0032 & 0.6332\(\pm\)0.0035 & 0.6214\(\pm\)0.0007  \\
COCO & 2$nd$    & 0.6035\(\pm\)0.0033 & 0.6314\(\pm\)0.0039 & 0.6181\(\pm\)0.0039  \\
COCO & 3$rd$    & 0.6030\(\pm\)0.0035 & 0.6300\(\pm\)0.0041 & 0.6212\(\pm\)0.0057  \\
Fundus & 1$st$   & 0.8082\(\pm\)0.0063 & 0.8122\(\pm\)0.0054 & 0.8031\(\pm\)0.0045         \\
Fundus & 2$nd$   & 0.8083\(\pm\)0.0058 & 0.8121\(\pm\)0.0054 & 0.7904\(\pm\)0.0203\\
Fundus & 3$rd$   & 0.8086\(\pm\)0.0059 & 0.8120\(\pm\)0.0053 & 0.7778\(\pm\)0.0149     \\
COVID QU-Ex  & 1$st$   & 0.4930\(\pm\)0.0014 & 0.4950\(\pm\)0.0007 & 0.4937\(\pm\)0.0001   \\
COVID QU-Ex  & 2$nd$  & 0.4929\(\pm\)0.0012 & 0.4951\(\pm\)0.0007 & 0.4879\(\pm\)0.0020  \\
COVID QU-Ex  & 3$rd$  & 0.4921\(\pm\)0.0014 & 0.4949\(\pm\)0.0005 & 0.4921\(\pm\)0.0002  \\
ISIC2018  & 1$st$    & 0.6984\(\pm\)0.0015 & 0.6670\(\pm\)0.0042 & 0.7367\(\pm\)0.0045 \\
ISIC2018  & 2$nd$    & 0.6992\(\pm\)0.0037 & 0.6683\(\pm\)0.0071 & 0.7483\(\pm\)0.0060 \\
ISIC2018  & 3$rd$    & 0.6983\(\pm\)0.0048 & 0.6682\(\pm\)0.0072 & 0.7436\(\pm\)0.0009  \\
\noalign{\smallskip}
\hline
\end{tabular}
\end{table*}

\section{Experiments on Additional Point Prompts}
As shown in Table \ref{tab:appendix_SAM_Point_Num}, the rows indicated by "3 points" in the "Points Number" column represent the SAMAug performance with two point prompts added to the initial point prompt. The selection of these two points followed the approach outlined in the main text. It can be observed that performing prompt augmentation will improve SAM's initial results across all four datasets. Specifically, the Max Distance method achieves the highest scores of 0.6473 and 0.7895 on COCO and Fundus datasets. The Random method achieves the highest score of 0.4996 on the COVID QU-Ex dataset. The Saliency method achieves the highest score of 0.74 on the ISIC2018 dataset. However, it is worth noting that, except for the Initial and Saliency results on the ISIC2018 dataset, which slightly outperform the two-point prompt, all other three-point prompts show a slight decrease in performance. To further validate the negative impact of increasing the number of point prompts, we also experimented with four additional point prompts (in the rows indicated by "5 points") and observed a similar trend. In summary, the experimental results indicate that increasing the number of augmented point prompts will not improve the overall segmentation performance. 

\section{Stability Study of the Point Prompt Selection on Model Performance}
To further validate the performance and stability of the two-point prompt, we selected three candidate points (rather than one) as the augmented point prompt for the Max Entropy, Max Distance, and Saliency sampling. Specifically, for Max Entropy, we selected the top three points with the highest cross-entropy to the initial point prompt and conducted three segmentation experiments accordingly. Similarly, in the Max Distance method, we selected the top three points based on their distance to the initial point prompt. In the Saliency method, we randomly selected three points from the saliency mask. To expedite the experiment, we randomly selected a subset of samples from each dataset. In COCO, we randomly selected 20 images per category, and for the other three medical datasets, we randomly selected 300 images each for the experiment.

The results are presented in Table \ref{tab:appendix_SAM_Point_Stability}. In the second column of Table \ref{tab:appendix_SAM_Point_Stability}, 1$st$, 2$nd$, and 3$rd$ represent the segmentation performance using each point prompts. It can be observed that the segmentation results are very consistent across the three point prompts across all datasets for the Max Entropy and Max Distance sampling, indicating that the points sampled using the same strategy will lead to similar SAM performance. For the Saliency method, slight fluctuations can be observed in the three results as the points were randomly selected from the saliency mask.

\end{appendices}

\end{document}